**Short Paper**

# Smart Face Shield: A Sensor-Based Wearable Face Shield Utilizing Computer Vision Algorithms

Manuel Luis C. Delos Santos
Asian Institute of Computer Studies, Quezon City, Philippines
manuelluisconchadelossantos@gmail.com
ORCID: 0000-0002-6480-3377
(corresponding author)

Ronaldo S. Tinio
Pamantasan ng Lungsod ng Valenzuela, Philippines
ORCID: 0000-0002-3529-8033

Darwin B. Diaz
Asian Institute of Computer Studies, Quezon City, Philippines

Karlene Emily I. Tolosa
Asian Institute of Computer Studies, Quezon City, Philippines

**Abstract**

*Purpose* – The study aims the development of a wearable device to combat the onslaught of covid-19. Likewise, to enhance the regular face shield available in the market. Furthermore, to raise awareness of the health and safety protocols initiated by the government and its affiliates in the enforcement of social distancing with the integration of computer vision algorithms.

*Method* – The wearable device was composed of various hardware and software components such as a transparent polycarbonate face shield, microprocessor, sensors, camera, thin-film transistor on-screen display, jumper wires, power bank, and python programming language. The algorithm incorporated in the study was object detection under computer vision machine learning. The front camera with OpenCV technology determines the distance of a person in front of the user. Utilizing TensorFlow, the target object identifies and detects the image or live feed to get its bounding boxes. The focal length lens requires the determination of the distance from the camera to the target object. To get the focal length, multiply the pixel width by the known distance and divide it by the known width (Rosebrock, 2020). The deployment of unit testing ensures that the parameters are valid in terms of design and specifications.

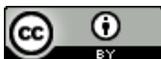

*Results* – With the application of the camera-to-distance to object mathematical concept and tag reference model, the prototype was subjected to unit testing using live feed scenarios. The distance between individuals was measured using several markers drawn on the floor to determine whether it falls within the safe, warning, and unsafe tag reference indicated by color schemes such as green, orange, and red with corresponding conditions. The device also tested its capability to automatically detect the measurement of the object in terms of height and compare its true value against the experimental value by using the percent error formula. The analysis revealed that the experimental values were closer to the acceptable or real values of less than 1.0-meter.

*Conclusion* – The technical design and unit testing validated the mathematical concepts, specifications, functionality, and machine learning algorithm. The percentage errors yielded a minor error indicating that the detected values were close to the actual or original values predicted in real-time. The SFS device has proven its potential use to become a timely innovation to help lessen the spread of the deadly covid-19 as an alternative to the traditional face mask and face shield.

*Recommendations* – For future modifications, an integration of a wireless personal thermal scanner would be a great advantage.

*Practical Implications* – The innovations integrated with the wearable smart-face shield can help users observe social distancing to combat the covid-19. Due to its expensive production costs, producing this kind of device for public consumption is not practical.

***Keywords***: Algorithm, Face Shield, Computer Vision, Wearable

## INTRODUCTION

The year 2020 has been very crucial since the world is in a global pandemic. According to World Health Organization (WHO), COVID-19 is an infectious disease caused by a new strain of coronavirus previously unidentified before the outbreak began in Wuhan, China, on December 2019. The virus was officially named last February 12, 2020. Later on January 30, 2020, the Philippines' Department of Health (DOH) reported the first case of COVID-19 on a 38-year-old female Chinese national. The first local transmission of the virus was confirmed (Halili, 2020) on March 7, 2020.

The Department of Health (2020), explains that coronaviruses are a large family of viruses causing a range of illnesses, from the common cold to infections such as those caused by Middle East Respiratory Syndrome-related Coronavirus (MERS-CoV) and Severe Acute Respiratory Syndrome-related Coronavirus (SARS-CoV).

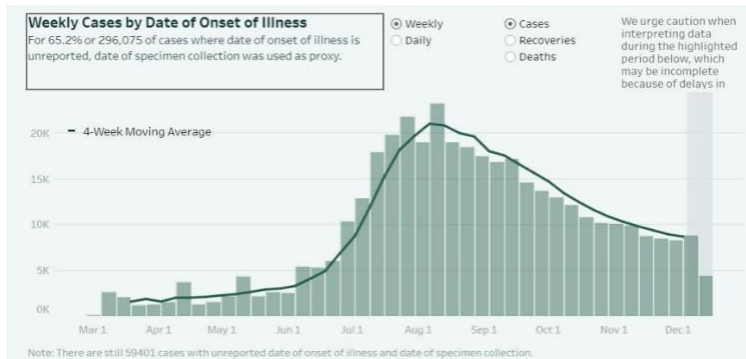

*Figure 1.* COVID-19 Case Tracker March 2020 to December 2020 from DOH

Based on Figure 1, the highest case was in August 2020 and gradually decreases during the second week of the same month up to December 2020.

A report from interaksyon.philstar.com, a photo taken by Edd Gumban, a photographer for a newspaper in the Philippines, shows commuters waiting for public transport in Metro Manila amidst the COVID-19 outbreak were non-compliance with social distancing protocols. Some people were not wearing their face masks and face shields properly (Melasig, 2020).

The top region that displayed the highest case of infected people were the area of NCR compared to the CALABARZON region (Calamba, Laguna, Batangas, Rizal, Quezon) and other regions in Central Luzon, Central Visayas, and Western Visayas. A dense population of workers was in Metro Manila.

For the protection of everyone against the spread of the virus, people acquaint themselves with the new normal. Those who were infected but did not show symptoms were primary factors in spreading the disease. These are the preventive measures people must practice (Centers for Disease Control and Prevention, 2020):

1. Maintain 1-meter distance from one another
2. Wear a face mask and face shield properly
3. Wash hands from time to time, or as needed
4. Maintain a proper hygiene
5. Avoid touching eyes, nose, and mouth
6. Cover the nose and mouth when sneezing or coughing
7. Clean and disinfect surfaces frequently

Authorities in the Philippines have implemented social distancing and preventive measures to protect every individual. Mass gatherings are prohibited. It has mandated to remain a 1-meter social distance requirement depicted in Figure 2. However, people, on the other hand, were negligent in conforming to the said implementations.

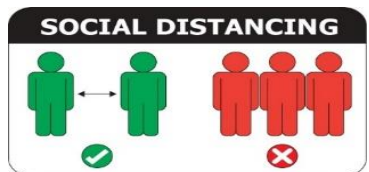

*Figure 2.* Social Distancing Representation

WHO encouraged people to be aware of different settings with different levels of risk as crowded places, close contact, and confined and enclosed spaces. The danger is higher in areas where these factors overlap (World Health Organization, 2020).

The Centers for Diseases Control and Prevention (CDC) is the national public health agency of the United States and defines social distancing as a safe space between an individual and other people that are not in the same household. The spread of the virus occurs when an infected person cough, sneezes, or talks, and the droplets of the saliva that mostly unnoticed are a risk to human transmission under the worst-case scenario, and these droplets could be inhaled and can reach the lungs.

Recognizing the rapidly evolving situation of the pandemic, this study aimed in the development of a smart, wearable device to enhance the regular face shield available in the market and raise awareness of the health and safety protocols initiated by the government and its affiliates in the enforcement of social distancing with the integration of computer vision algorithm.

**Objectives of the Study**

The main focus was to develop a science-based approach to determining the possible and available help to lessen the spread of the virus with specific objectives as follows:

1. Design and develop the prototype that would enhance the regular face shield that is already available on the market.
2. Integrate hardware and software components with appropriate computer vision algorithms for object detection.
3. Apply mathematical concepts to measure safe, warning, and unsafe distance for social distancing mechanisms using the camera to distance formula and tag reference model.
4. Test the prototype through unit testing to validate its functionality in terms of distance zone between people within the safe, warning, and unsafe range, compute for percentage error of the actual distance against detected distance and predict height in real-time.

## LITERATURE REVIEW

Several research papers and textbooks in high-level mathematics and statistics were used to define and explore various mathematical concepts, including the statistical metrics for the experimental design to be used for testing.

Scalinci and Battagliola (2020) stressed that when tiny droplets touch mucosal surfaces, such as the lips, nose, or eyes, the transmission of the virus occurs, and when someone sneezes, coughs, or speaks, tiny droplets in the air are released.

Alvarez et al., (2020) clarified that coronavirus disease 19, or covid-19 is a respiratory illness that causes severe pneumonia in an infected person. Acquiring this disease is through direct contact with generated respiratory droplets, saliva, or discharge from the nose when the infected person coughs or sneezes or through breathing in the virus if you are within the range of an infected person.

Cuenca et al., (2020) reiterated that stricter social distancing involves containment strategies. Works are scheduled properly to retain social distance and work from home with such job descriptions. The virus tends to be slowed by social isolation and quarantine in age-specific communities. As a result, interventions for these age groups and their households could be appropriate.

Hayes (2020) defined augmented reality (AR) as a technologically improved version of the natural world that is achieved by the use of artificial visual objects, vibration or sound, or other sensory inputs. It's a growing movement in businesses that deal with mobile computing and business applications.

Punn et al., (2020) recommended real-time-based deep learning to monitor social distancing using object detection and tracking approaches. The number of violations was given by computing the number of groups formed. The violation index term calculates the ratio of the number of people to the number of groups.

Ghorai et al., (2020) proposed a deep learning solution that would alert the person as soon as one violates social distancing. A video stream captures from the CCTV camera, and with the PoseNet model, the people are detected and then kept a rack of the number of people present in the video stream. If the distance between 2 frames of people is less than the prescribed social distance, the authorities in charge are then alerted.

Sai and Sasikala (2019) stated that object detection utilizes in several applications such as detecting vehicles, face detection, autonomous vehicles, and pedestrians on streets.

TensorFlow's Object Detection API is a powerful tool that can quickly enable anyone to build and deploy powerful image recognition software. Object detection not solely includes classifying and recognizing objects in an image but localizes those objects and attracts bounding boxes around them.

Keniya and Mehendale (2020) offered an answer to determine whether or not an individual is following the rule of social distancing. The findings are verified using a live stream and a video feed. By measuring the gap between two frames of people from the centroids, they can understand whether or not a person is maintaining social distancing. Also, they are labeled as safe and unsafe, as depicted in Figure 3.

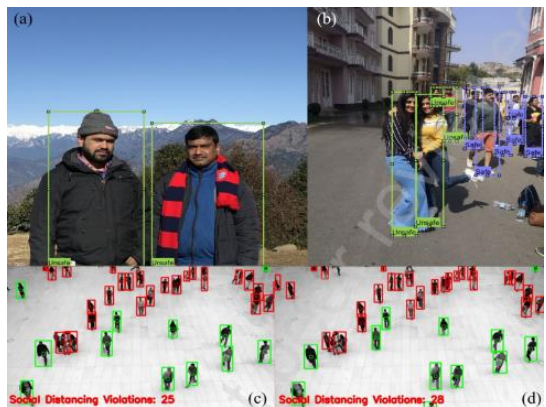

*Figure 3*. Image and Live Feed Video Input Scenarios

Figure 3 describes the scenarios as videos and images given to the model as inputs. In (a) and (b), people got detected as maintaining social distancing or not, depending on the distance maintained between two individuals. People marked in frames with different colors as green for violating social distancing and labeled as unsafe, and purple frame marked for maintaining social distancing and labeled as safe. In (c) and (d), the people got detected and marked as per the distance between two individuals. Along with this, the number of violations counts.

Ellims et al., (2004) explained that unit testing is a technique that receives a lot of criticism in terms of the amount of time it is perceived to take and how much it costs to perform. However, it is also the most effective mean to test individual software components for boundary value behavior and ensure that codes are adequately exercised.

## METHODOLOGY

The study aimed to provide a wearable device that uses supplementary face protection for social distancing. Primarily, the hardware components were composed of poly-carbonate transparent face shield masks. An augmented reality (AR) on-screen display via 5 inches TFT screen, Adafruit AMG8833 infrared thermal sensor, a version 2 Pi camera, an active-passive infrared and ultrasonic sensors for motion detection wired up to Raspberry

Pi version 4, and powered by a power bank as shown in Figure 4. The software components used Python programming language for TensorFlow Lite.

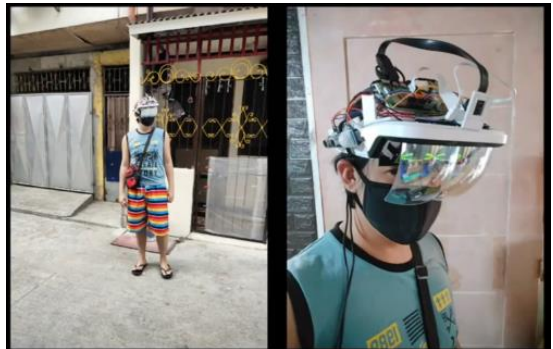

*Figure 4.* The Wearable Smart Face Shield (SFS) Prototype

Figure 5 shows the pictorial representation of the wearable smart face shield in a block diagram. The block consists of all Input (cameras and sensors), Processes (microprocessor, OpenCV, TF Lite, and GUI), and Output (LCD and AR Displays) components.

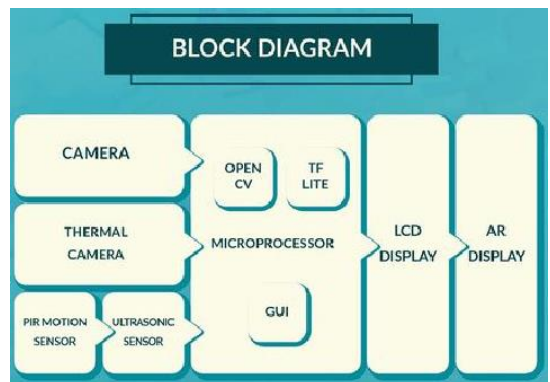

*Figure 5.* SFS Block Diagram

**Object Detection Algorithm**

The algorithm utilized computer vision's object detection using TensorFlow which is the process of acquiring data, training models, serving predictions, and refining future results (Chatterjee, 2020).

The front camera with OpenCV technology determines the distance of the object referred to as "person" (single individual) or "people" (more than one individual) in front of the SFS user.

As seen in Figure 5, the camera-to-distance flowchart depicts that when a person is detected on the right, left, and back sides, the distance zones of the user will trigger the motion sensor. The ultrasonic sensor activates and takes the distance measurement of the person who entered the distance zone. Refer to Table 1 for the condition applied to the SFS tag reference model.

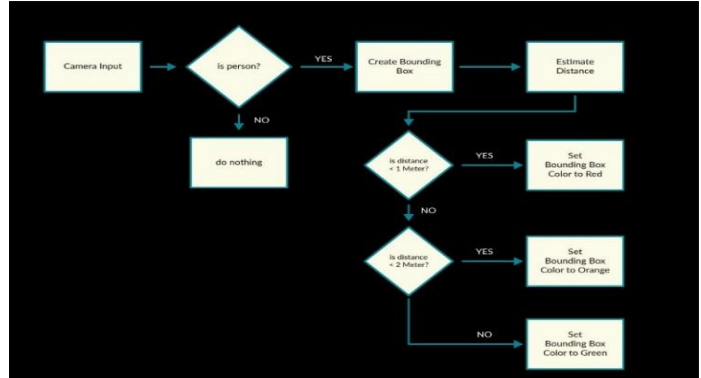

*Figure 5.* Camera-to-Distance Flowchart

Using TensorFlow's object detection algorithm, the target object identifies and processes the image or live video feed to get its bounding boxes shown in Figure 6.

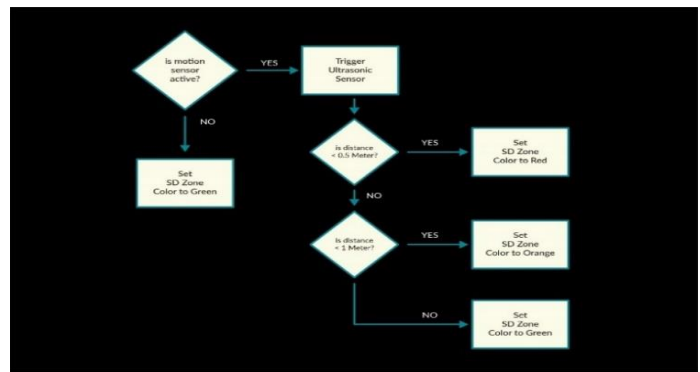

*Figure 6.* Object Detection Flowchart

**Graphical User Interface**

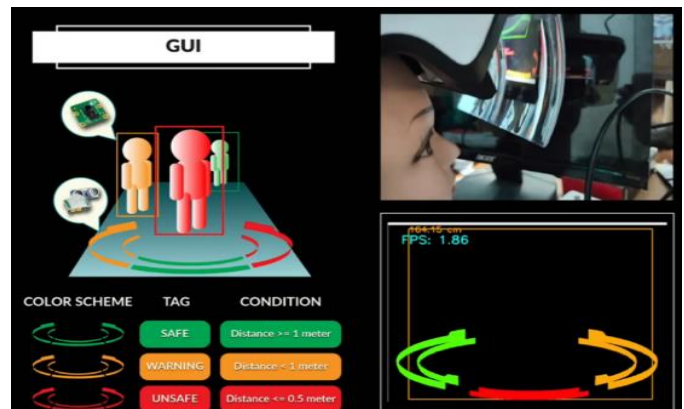

*Figure 7.* The Graphical User Interface

Figures 8 and 9 describe distance zones, color schemes, conditions, and the on-screen display of the GUI.

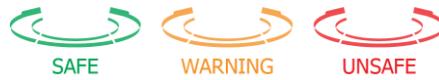

*Figure 8.* Distance Zones Tag Reference with Color Schemes

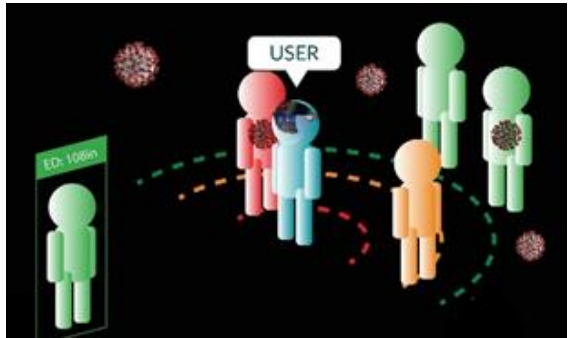

*Figure 9.* Distance Zone Representation

The user of the SFS within the distance zone changes color according to the condition of the person as explained in Table 1.

Table 1. Distance Zone According to Tag Reference, Conditions, and Color Schemes

| Tag Reference | Condition | Color Scheme |
|---|---|---|
| Safe | Distance >= 1.0 meter | Green |
| Warning | Distance <= 0.999 meter | Orange |
| Unsafe | Distance <= 0.5 meter | Red |

The 1.0-meter distance radius referred to the sector-relevant guidelines of the Inter-Agency Task Force (IATF), which oversees the health and safety protocols of the Philippine government, such as social distancing (Official Gazette, 2020) and the Department of Health (DOH) under Administrative Order No. 2020-0015 series of 2020.

**Camera-to-Distance Measurement**

To determine the distance from the camera to the target object. A calculation of the focal length of the lens of the camera is needed. To get the focal length, multiply the pixel width by the known distance. Divide it by the known width (Rosebrock, 2020), defined in equation 1.

$$\boldsymbol{F = (P * D)/W} \qquad \text{Equation 1}$$

Where:
**F** - Focal length, **P** - Pixel Width, **D** - Known Distance, **W** - Known Width

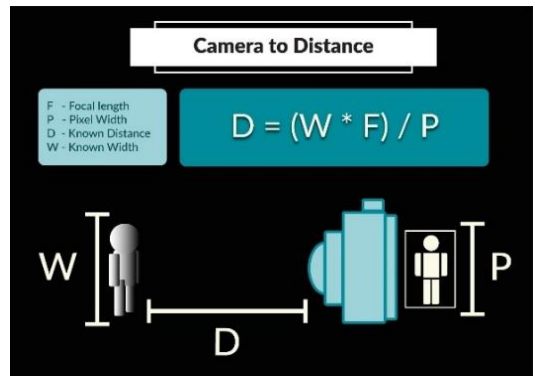
*Figure 10*. Camera-to-Distance to Object

Figure 10 clarifies the camera-to-distance to object using the triangle similarity method, as the measured distance is equal to the known width, multiplied by the focal length of the lens of a camera over pixel width as defined in equation 2. To determine the distance from the camera to the target object. A calculation of the focal length for the camera is needed as defined in equation 2.

$$\boldsymbol{D = (W * F)/P} \qquad \text{Equation 2}$$

In this case, the known width is the average Filipino height which is 5'4" (Punongbayan, 2020). The pixel width is the height of the bounding box.

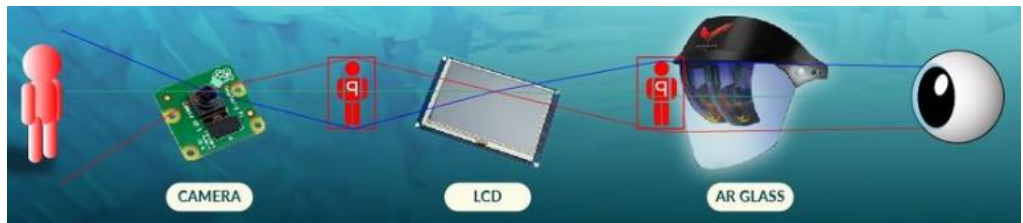
*Figure 11*. The Augment Reality Display Representation

Figure 11 illustrates the flow of input from the sensors, and camera, which is then displayed on the screen and reflected in the AR glass. The information is overlaid in real-time in the real world.

**Percent Error**

Percent errors express how big errors are as you measure something in an experiment. Smaller values mean that you are close to the accepted or real value. The accepted value is sometimes called the "true" value or "theoretical" value (Glen, 2021) shown in equation 3.

**PE = (|experimental value - true value| / experimental value) x 100**         *Equation 3*

**Unit Testing**

The device was subjected to unit testing to validate its functionality with different parameters. The importance of unit testing plays a key role and was used in building high-quality software practice (Gren & Antinyan, 2017).

## RESULTS AND DISCUSSION

With the application of the camera-to-distance to object mathematical concept and tag reference model, the prototype was subjected to unit testing using live feed scenarios. The actual distance measurements of two people with several markers were drawn on the floor as shown in Figure 12, such as less than or equal to 1.0 meter for the "Safe" tag reference, less than 1.0 meter for the "Warning" tag reference and less than 0.5 meters for the "Unsafe" tag reference.

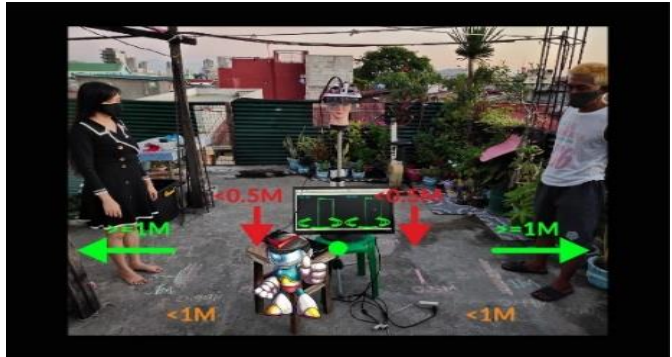

*Figure 12.* Testing distance zones with marked distances

As people entered near the SFS user's distance zone, as demonstrated in Figure 13, two people were standing on opposite sides. The graphical user interface in the distance zone changed its color to "Green" based on the conditions measured by the ultrasonic sensor, which is greater than or equal to 1.0 meters.

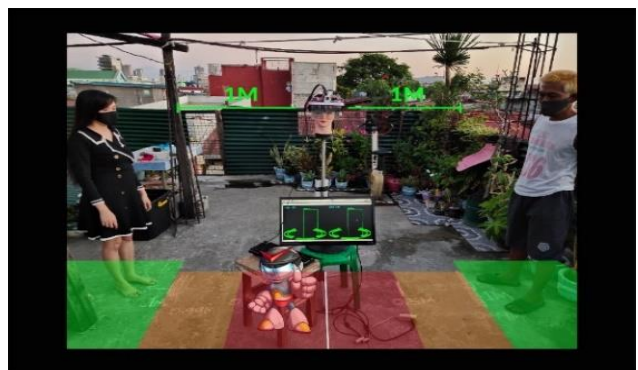

*Figure 13.* Testing Range – Safe Zone

While the person on the left side in Figure 14 moved toward the "Warning" distance zone as indicated by the "Orange" color scheme, the conditions between the two people were less than or equal to 0.999 meters. As observed, the person standing on the right side was in the "Safe" distance zone, which fell under the "Green" color scheme.

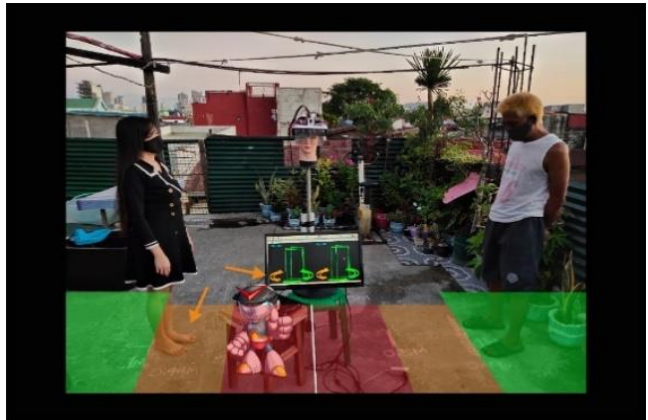

*Figure 14.* Testing Range - Warning

When the person on the left side in Figure 15 moved toward the "Warning" distance zone within less than or equal to 0.5-meter conditions, the person entered the "Unsafe" distance zone, which is indicated by the "Red" color scheme.

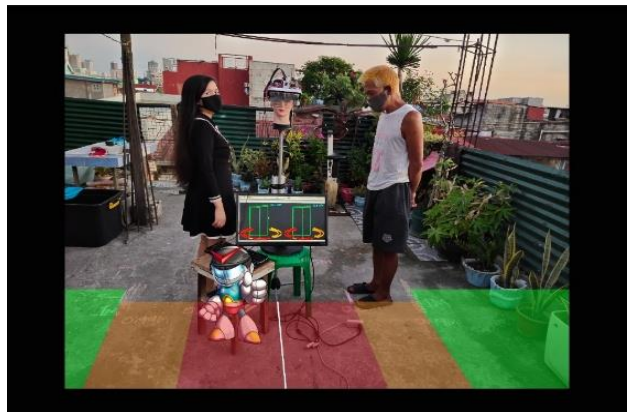

*Figure 15.* Testing Range - Unsafe

Figure 16 compares the actual distance versus the estimated distance detected by the device. In the three consecutive scenarios, the estimated distance of 2.02 meters (right side), 2.95 meters (middle side), and 4.06 meters (left side) versus the actual distance as indicated by markers of 2.0 meters (right side), 3.0 meters (middle side) and 4.0 meters (left side) with the difference of 0.02 meter, -0.05 meter, and 0.06 meter and percent errors of 0.99%, 1.70% (disregard negative values), and 1.48% respectively by applying the formula in equation 4.

**PE = Detected Distance – Actual Distance / Detected Distance x 100**               *Equation 4*

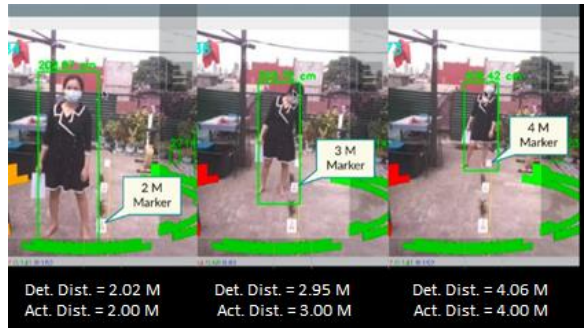
*Figure 16*. Detected Distance vs. Actual Distance

Figure 17 detects the actual height of a person in inches, which is indicated by the inbound box as 141.00 inches and validated by the actual use of a measuring tape similar to 141.00 inches.

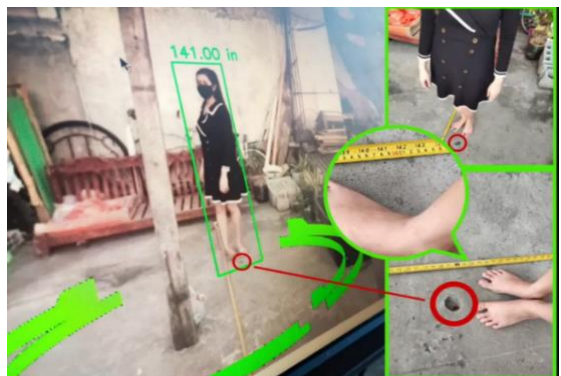
*Figure 17*. SFS height detection is measured in inches

**LIMITATIONS**

The device does not store datasets due to its limited storage capacity and uses a live feed for image processing and object detection. It can only detect and measure an object's distance within the frontal camera view. The measurement accuracy depends on the height of the person since the effective range of the ultrasonic sensor used in the study was around 4.0 meters only. However, using a much higher effective range ultrasonic sensor rectifies the issue.

**CONCLUSIONS AND RECOMMENDATIONS**

The technical design and unit testing validate the mathematical concepts, specifications, functionality, and machine learning algorithm. The percentage errors yielded a minor error indicating that the detected values were close to the actual or original values predicted in real-time. The SFS device has proven its potential use to become a timely innovation to help lessen the spread of the deadly covid-19 as an alternative to the traditional face mask and face shield.

For future modifications, an integration of a wireless thermal scanner would be a great advantage.

## PRACTICAL IMPLICATIONS

The innovations integrated with the wearable smart-face shield can help users observe social distancing to combat the covid-19. Due to its expensive production costs, producing this kind of device for public consumption is not practical.

## ACKNOWLEDGEMENT

The researchers would like to extend their sincerest gratitude to the authors of various works of literature used in this study. Likewise, to the Asian Institute of Computer Studies, and all the unselfish people who made the publication of this study possible despite the threat of Covid-19.

## DECLARATIONS

**Conflict of Interest**

All authors declared that they have no conflicts of interest.

**Informed Consent**

All participants were appropriately informed and voluntarily agreed to the terms with full consent before taking part in the conduct of the experiment.

**Ethics Approval**

The AICS Research Ethics Committee duly approved this study on November 2020 after it conformed to the local and international accepted ethical guidelines.